\documentclass[letterpaper, 10 pt, conference]{ieeeconf}  

\IEEEoverridecommandlockouts                              

\usepackage{listings}
\usepackage[pagebackref,breaklinks,colorlinks]{hyperref}
\usepackage{graphicx} 
\usepackage[noend]{algpseudocode}
\usepackage{algorithmicx}
\usepackage{algorithm}
\usepackage{amssymb, amsmath, amsfonts}
\usepackage{booktabs}
\usepackage{bbm}
\usepackage{booktabs}
\usepackage{multirow}
\usepackage{colortbl}
\usepackage{amssymb} 
\usepackage{pifont} 
\usepackage{subcaption}
\usepackage{tabularx}

\MakeRobust{\Call}

\usepackage{url}
\usepackage{cleveref}
\usepackage[sort]{cite}

\usepackage[table]{xcolor}
\definecolor{mygray}{gray}{.9}

\newcommand{\methodname}{NEUSIS}
\newcommand{\methodfull}{Neuro-Symbolic Intelligent Search}
\newcommand{\methodfullHighlight}{\textbf{Neu}ro-\textbf{S}ymbolic \textbf{I}ntelligent \textbf{S}earch}
\newcommand{\perceptionname}{GRiD}
\newcommand{\perceptionfull}{Perception, Grounding, Reasoning in 3D}

\newcommand{\manoeuvername}{SNaC}
\newcommand{\manoeuverfull}{Selection, Navigation and Coverage}

\usepackage{soul}

\usepackage{listings}

\definecolor{codegreen}{rgb}{0,0.6,0}
\definecolor{codegray}{rgb}{0.5,0.5,0.5}
\definecolor{codepurple}{rgb}{0.58,0,0.82}
\definecolor{backcolor}{rgb}{0.95,0.95,0.92}

\lstdefinestyle{code}{
    backgroundcolor=\color{backcolor},   
    commentstyle=\color{codegreen},
    keywordstyle=\color{magenta},
    numberstyle=\tiny\color{codegray},
    stringstyle=\color{codepurple},
    basicstyle=\ttfamily\footnotesize,
    breakatwhitespace=false,         
    breaklines=true,                 
    captionpos=b,                    
    keepspaces=true,                 
    numbers=left,                    
    numbersep=5pt,                  
    showspaces=false,                
    showstringspaces=false,
    showtabs=false,                  
    tabsize=2,
    columns=fullflexible,
}

\title{NEUSIS: A Compositional Neuro-Symbolic Framework for Autonomous Perception, Reasoning, and Planning in Complex UAV Search Missions}

\author{Zhixi Cai$^{*}$, Cristian Rojas Cardenas$^{*}$, Kevin Leo$^{*}$, Chenyuan Zhang$^{*}$, Kal Backman$^{*}$, Hanbing Li$^{*}$, \\ Boying Li, Mahsa Ghorbanali, Stavya Datta, Lizhen Qu, Julian Gutierrez Santiago, Alexey Ignatiev,\\ Yuan-Fang Li$^\text{†}$, Mor Vered$^\text{†}$, Peter J. Stuckey$^\text{†}$, Maria Garcia de la Banda$^\text{†}$ and Hamid Rezatofighi$^\text{†}$
\thanks{*These authors contributed equally to this work as the joint first authors}
\thanks{†These authors contributed equally to this work as the joint last authors}
\thanks{All authors are with Faculty of Information Technology,
        Monash University, 3800 VIC, Australia
        {\tt\small firstname.surname@monash.edu}}%
\thanks{This work is supported by the DARPA Assured Neuro Symbolic Learning and Reasoning (ANSR) program under award number FA8750-23-2-1016}%
}

\begin{document}
\maketitle
\thispagestyle{empty}
\pagestyle{empty}


\begin{abstract}

This paper addresses the problem of autonomous UAV search missions, where a UAV must locate specific Entities of Interest~(EOIs) within a time limit, based on brief descriptions in large, hazard-prone environments with keep-out zones. The UAV must perceive, reason, and make decisions with limited and uncertain information. 
We propose \methodname{}, a compositional neuro-symbolic system designed for interpretable UAV search and navigation in realistic scenarios. \methodname{} integrates neuro-symbolic visual perception, reasoning, and grounding (\perceptionname{}) to process raw sensory inputs, maintains a probabilistic world model for environment representation, and uses a hierarchical planning component (\manoeuvername{}) for efficient path planning.
Experimental results from simulated urban search missions using AirSim and Unreal Engine show that \methodname{} outperforms a state-of-the-art~(SOTA) vision-language model and a SOTA search planning model in success rate, search efficiency, and 3D localization. These results demonstrate the effectiveness of our compositional neuro-symbolic approach in handling complex, real-world scenarios, making it a promising solution for autonomous UAV systems in search missions.

\end{abstract}

\section{Introduction}
\label{sec:Intro}

The development of autonomous agents capable of safely completing Intelligence, Surveillance, and Reconnaissance~(ISR) missions in complex environments presents significant challenges~\cite{firooziFoundation2023}.
Unmanned Aerial Vehicles~(UAVs) are increasingly utilized in these missions due to their ability to cover large areas and access hazardous locations with minimal risk to human life~\cite{MOHDDAUD202230}. However, designing fully autonomous UAV systems for such tasks, given onboard sensory data and brief mission descriptions in unpredictable and complex environments with uncertain knowledge, remains a formidable challenge.

In this paper, we focus on life-like scenarios in which a UAV must, within a designated time limit, autonomously search for a number of specific Entities of Interest~(EOIs) based on brief descriptions, e.g., find ``\emph{a red SUV vehicle}'' or ``\emph{a pedestrian carrying a blue umbrella}'', in a large suburban or urban environment that may contain hazards or keep-out zones~(KOZs). These hazard zones represent significant risks that the UAV must carefully avoid while efficiently searching designated Areas of Interest~(AOIs)~\cite{risk-aware}. 
To successfully operate in such scenarios, an autonomous UAV must actively and reliably perceive the environment from onboard sensory measurements, reason about the environment, and make decisions based on the mission description and partial or uncertain information about the surroundings.

\begin{figure}[t]
    \centering
    \includegraphics[width=\linewidth]{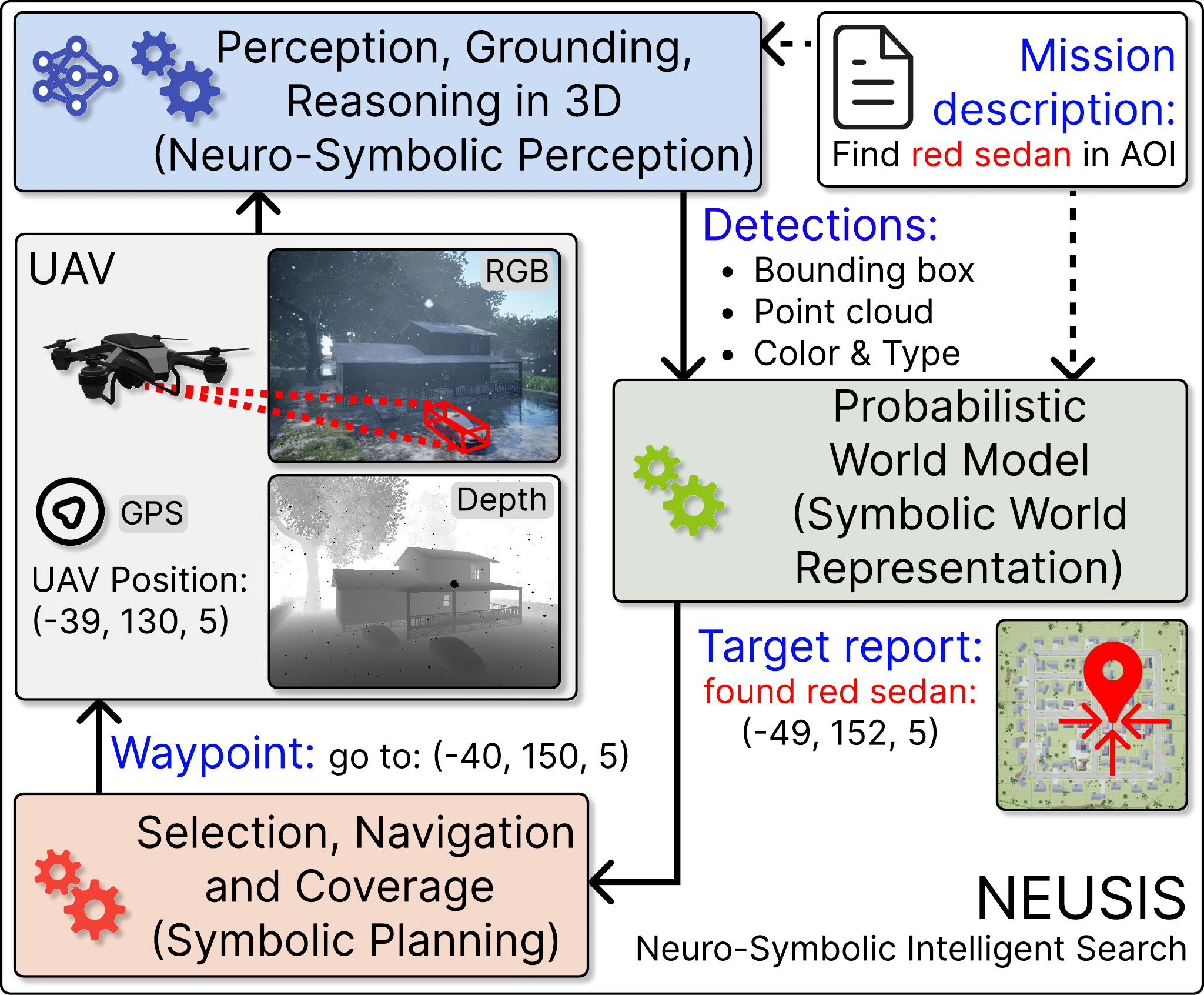}
    \caption{Overview of \methodname{}. Neuro-symbolic \textit{\perceptionfull{}} (\perceptionname{}); Symbolic Probabilistic World Model; and \textit{\manoeuverfull{}} (\manoeuvername{}) components autonomously complete UAV search missions by processing sensor inputs to find targets, such as the red sedan required by the mission description.}
    \label{fig:teaser}
    \vspace{-2mm}
\end{figure}

\begin{figure*}[t]
    \centering
    \includegraphics[width=\linewidth]{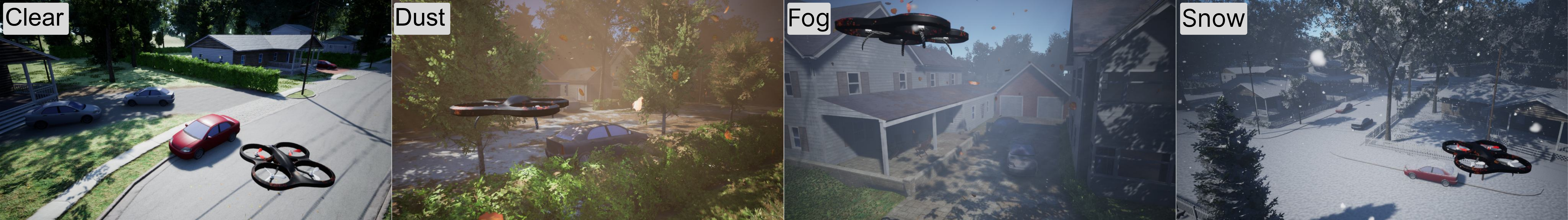}
    \caption{Screenshots from the Neighborhood environment illustrating different real-world challenges for UAVs.}
    \label{fig:environment}
    \vspace{-2mm}
\end{figure*}

Recently, advances in Large Multimodal Models (LMMs)~\cite{shahLMNav2023,shahViNT2023,paloUnified2023} have shown promise in different robotics tasks. However, their reliance on diverse, large-scale datasets for training as monolithic end-to-end models imposes significant computational demands. These models often lack interpretability when they fail and struggle to generalize beyond their training domains, especially in adversarial settings~\cite{huangLearning2021,sunNeuroSymbolic2021}. Furthermore, LMMs lack explicit components to model the world state or update their knowledge, which is crucial for complex tasks like searching in unconstrained environments.
Alternatively, many autonomous robotics systems employ compositional approaches~\cite{linText2Motion2023,brohanRT22023,mahmoudiehZeroShot2022}, integrating explicit perception and planning to perform tasks. These systems use neural-based perception models to process sensory data into abstract representations like segmentation, detection, or captions, which a neural planner then uses for navigation. While these approaches offer better interpretability and generalizability than monolithic models, they still lack explicit visual reasoning and world state representation. Neural planners require substantial training data, are task-constrained, and may be less efficient than model-based or symbolic planners. They also remain vulnerable to adversarial conditions, making them less suitable for search problems in unconstrained environments, the focus of this paper.

A viable baseline for such search problems is to use state-of-the-art~(SoTA) vision or multimodal language models~\cite{cheng2024yolo, liuGrounding2023, li2022grounded} to process mission specifications and sensory data, projecting it to a robust abstract level. This can then be integrated with model-based planners~\cite{Mier_Fields2Cover_An_open-source_2023, macenski2020marathon2, yang2014literature, shahViNT2023} that offer improved generalizability, robustness, and efficiency.
However, this approach still lacks explicit visual reasoning and a persistent world model, which limits the ability to maintain an interpretable representation of the environment and make informed decisions.

To address these limitations, we introduce \methodname{} (\methodfullHighlight{}), a novel compositional neuro-symbolic framework comprised of three main components (see Figure~\ref{fig:teaser}): 
\emph{(i)} a Neuro-symbolic component for \emph{Perception, Grounding and Reasoning in 3D}, \perceptionname{}, which handles the perception, visual reasoning and localization of entities of interest in a 3D world using UAV visual sensors;
\emph{(ii)} a \emph{Probabilistic (Symbolic) World Model}, which refines the potentially noisy outputs from \perceptionname{} and updates the belief about entities of interest based on world knowledge and probabilistic belief, maintaining a coherent and interpretable representation of the environment that enables robust reasoning and decision-making;
\emph{(iii)} a \emph{Hierarchical Model-Based (Symbolic) Planning} component, \manoeuvername{}, which uses high-level planning to determine the overall search strategy, mid-level planning for navigating to an allocated area, and low-level planning to efficiently and effectively search within allocated areas  while avoiding obstacles.

\begin{figure}[t]
\centering
\includegraphics[width=0.70\linewidth]{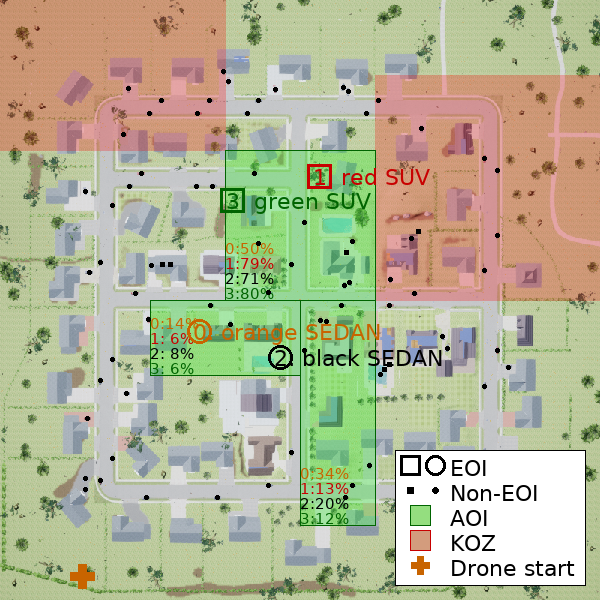}
\caption{An example mission scenario with four EOIs spread across three AOIs. 
Prior likelihood of EOI presence is shown in the bottom left corner of each AOI.
}
\label{fig:missionVisualisation}
\vspace{-2mm}
\end{figure}

We evaluate \methodname{} on a search mission benchmark developed by Keno~\emph{et al.}~\cite{kenoSimulationbased2024} based on the AirSim~\cite{shah2018airsim} simulation platform for Unreal Engine as part of the DARPA-Assured Neuro-symbolic Reasoning~(ANSR) program\footnote{\url{https://www.darpa.mil/program/assured-neuro-symbolic-learning-and-reasoning}}.
This benchmark presents complex scenarios, including various challenging environmental settings (e.g., different weather conditions). Our results demonstrate that \methodname{} significantly outperforms a strong compositional baseline in terms of success rate, navigation efficiency, and target localization, marking a significant advancement in end-to-end autonomous UAV systems.

\begin{figure*}[t]
    \centering
    \includegraphics[width=\linewidth]{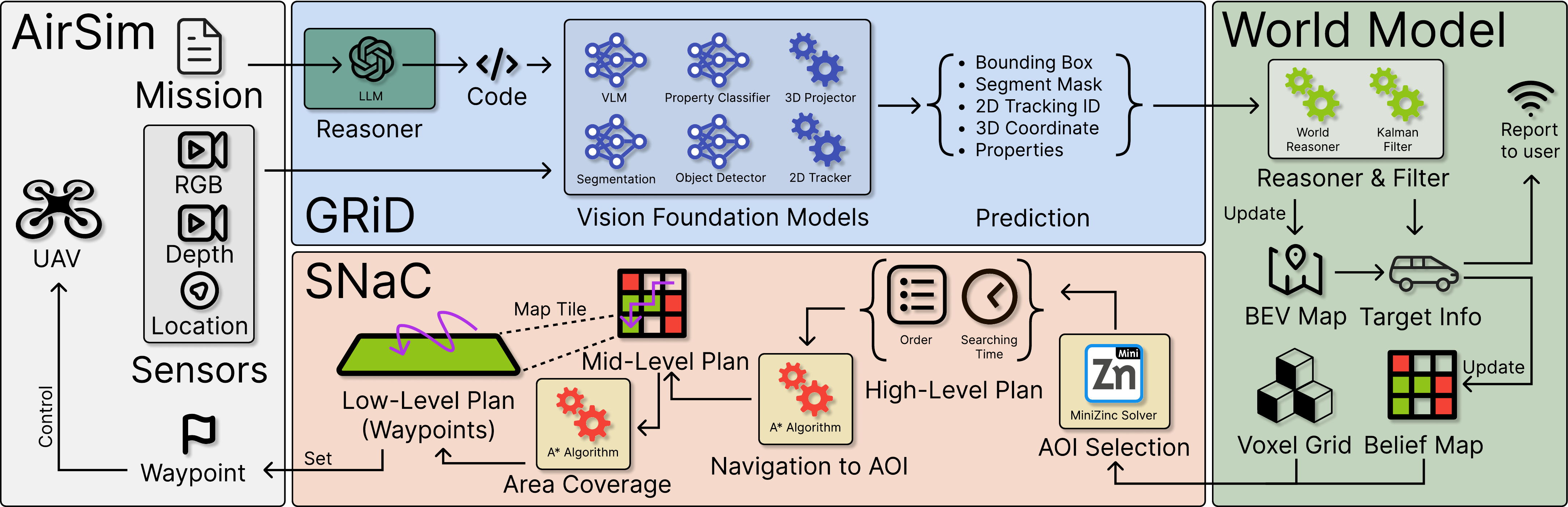}
    \caption{The pipeline of our proposed neuro-symbolic system, \methodname{}. The UAV operates in a simulated environment (AirSim) and is equipped with sensors including RGB camera, depth camera, and GPS. The \textit{\perceptionfull{}} (\perceptionname{}) component processes sensor data using a reasoner (code generator) and Vision Foundation Models (VFMs), including neuro-based segmentation, object detection, property classification, and symbolic 2D tracker and 3D projector, to generate predictions. Predictions are sent to the world model, which maintains a belief map, and generates target reports. The \textit{\manoeuverfull{}} (\manoeuvername{}) component generates a hierarchical plan, with the AOI Selection, AOI Navigation, and Area Coverage modules producing high-level, mid-level, and low-level plans.}
    \label{fig:pipeline}
    \vspace{-2mm}
\end{figure*}

\section{Environment and Mission} 
\label{sec:env_task}

Our benchmarks are based on the Hybrid AI Mission Environment for RapId Training and Testing~(HAMERITT) system presented in~\cite{kenoSimulationbased2024} for evaluating participants of the DARPA ANSR program.
HAMERITT is a platform for UAV simulation and testing, based on the AirSim~\cite{shah2018airsim} plugin for Unreal Engine, capable of dynamically generating complex evaluation scenarios.
In this section, we will provide a high-level overview of the features used for our benchmarks.
A detailed explanation of its capabilities can be found in~\cite{kenoSimulationbased2024}.

\subsection{Environment dataset}
We use HAMERITT's Neighborhood environment for emulating surveillance and reconnaissance missions in urban settings. 
The environment contains a broad 500x500~meter search area densely populated with a diverse selection of world objects (e.g., houses, trees, fences, roads, and non-EOI cars), that can be adversarially positioned to challenge perception with occlusions, and planning with complex navigation tasks. 
Figure \ref{fig:environment} show example images from this environment. 
Scenarios can also take place at night or include adverse weather conditions~(e.g., snow, fog, and leaves) that further challenge the clarity of sensor information.

\subsection{Mission}
The UAV's mission is to identify as many entities-of-interest~(EOI) as possible within the specified areas-of-interest~(AOIs) and time constraints. %
EOIs are specified using a combination of descriptions (e.g., ``red SUV'') and a probability for being  within different AOIs.
Further limitations are imposed via keep-out-zones~(KOZs) which denote areas the UAV must not enter. 
For our benchmarks a time limit of $5$~minutes is imposed. 
The environment is populated with cars that  have an associated type~(sedan or SUV), and color~(from $8$ potential colors). 
EOIs are chosen such that a unique description can be formulated that is not ambiguous within the proposed AOIs.
To be successful, the UAV must prioritize its focus on the most promising AOIs and allocate its time wisely. 
Visualization of a potential mission scenario is shown in Figure~\ref{fig:missionVisualisation}. 
To challenge robustness under adversarial conditions, distracting non-EOI cars that partially match an EOI description~(e.g., with the correct type and color, but positioned outside the AOI) may also be present.

\section{The \methodname{} System}\label{sec:system} 
\methodname{}, \methodfull{}, is a compositional framework comprised of three main components: a neuro-symbolic visual perception, reasoning and grounding component~(\perceptionname{}), a symbolic world model, and a symbolic hierarchical planning component~(\manoeuvername{}), shown in Figure~\ref{fig:pipeline}.

\subsection{\perceptionfull{}~(\perceptionname{})}
\label{sec:percept}

The \textit{\perceptionfull{}}~(\perceptionname{}) component processes UAV sensor data, i.e., RGB, depth, and GPS location for robust visual reasoning and 3D object grounding in complex search missions. A key challenge in the UAV missions described in Section~\ref{sec:env_task} is not only localizing EOIs in 3D space but also inferring their attributes. \perceptionname{} addresses this by integrating visual perception, grounding, and neuro-symbolic reasoning.

\perceptionname{} builds on recent advances in neuro-symbolic compositional visual reasoning methods~\cite{suris2023vipergpt, you2023idealgpt, lu2023chameleon, gupta2023visual, stanic2024towards}, which tackle complex visual reasoning tasks, such as visual grounding, by decomposing them into sub-tasks. These sub-tasks are individually solved using vision foundation models and large language model~(LLM)-generated code, with the 
results combined to complete the overall task. For \perceptionname{}, we adopt HYDRA~\cite{keHYDRA2024}, a state-of-the-art neuro-symbolic reasoning system that combines reinforcement learning with LLM-driven code generation to enable dynamic, compositional visual understanding. While HYDRA is designed for 2D image-based reasoning tasks (e.g., visual grounding and question answering), it requires adaptation to handle perception, reasoning, and grounding from the sensor stream of visual data in the UAV’s 3D search mission.

To adapt HYDRA for this mission, we expand \perceptionname{}’s toolkit to include 2D target bounding boxes with attributes recognition, instance segmentation, object tracking, and 3D coordinates projection. The following Python APIs are implemented to meet mission requirements: $\mathbf{segment}$, $\mathbf{classify\_object\_attributes}$, $\mathbf{classify\_object\_types}$, $\mathbf{track}$, and $\mathbf{project\_to\_3d}$.
We integrate state-of-the-art VFMs for grounding, segmentation, property classification, and 2D tracking. CLIP~\cite{radfordLearning2021} was fine-tuned for classifying object attributes and types ($\mathbf{classify\_object\_attributes}$, $\mathbf{classify\_object\_types}$), while pretrained models~\cite{liuGrounding2023, xiong2023efficientsam} were used for grounding ($\mathbf{find}$, $\mathbf{segment}$). Object tracking ($\mathbf{track}$) employed a symbolic method~\cite{cao2023observation}. Further implementation details are provided in Section~\ref{sec:implementation_details}.

To avoid computational bottlenecks and resource overuse, we implemented a caching mechanism for the LLM-generated reasoner code, allowing reuse of reasoner plans across similar mission queries in different scenarios.

\subsection{Probabilistic World Model} \label{sec:world_model}

Due to the inherent noise in sensor inputs, \perceptionname{} is rarely 100\% confident in its output. The world model accumulates localization information from \perceptionname{} to maintain a persistent probabilistic representation of the environment and provide a mechanism for identifying and reporting EOIs.

The world model is initialized with a ground-truth voxel occupancy grid and a birds-eye view~(BEV) segmentation map, indicating the locations of static objects like walls, trees, and roads.
It is also provided with the initial prior belief map from the mission description.
For each frame, it receives noisy 3D localization data and attributes from \perceptionname{} and performs the following tasks:

\subsubsection{World Reasoning} To refine 3D localization, the world model uses domain knowledge, such as removing infeasible points from masked depth data to compute more accurate 3D center points, and discarding detections that violate physical constraints (e.g., cars high above ground or inside walls).

\subsubsection{Information Accumulation} 
To improve 3D localization and attribute classification the world model can accumulate detections about the same objects over time.
A na\"ive approach would compute the average position of detections, but this may not take into account the uncertainty of \perceptionname{}'s outputs.
Instead, we use (i) Bayesian filtering for position refinement and (ii) discrete attribute distribution ranking updates for more accurate attribute likelihoods.
This process enhances the probabilistic model of the world, increasing confidence in positions and attributes over time.

\subsubsection{Reporting} The world model generates online reports by evaluating whether any tracked objects match the EOI descriptions. It reasons about the probability of a match, and any candidates exceeding a confidence threshold are reported. A final offline report summarizing the best detection of each EOI is produced at the end of the mission.

In addition to the functions mentioned above, the World Model also maintains  environmental information relevant to the planning component, including the voxel occupancy grid and belief map,  to support path planning operations.

\subsection{\manoeuverfull{} (\manoeuvername{})}
\label{sec:maneuver}

The planning component is designed to generate a trajectory that efficiently searches the AOIs by maximizing the likelihood of encountering EOIs within the allocated time. The component first retrieves the belief map and other environmental information from the World Model component, and then generates a sequence of waypoints to be sent to the UAV's control unit.
While this task is closely related to area coverage and object goal navigation, the existence of \textit{multiple} EOIs within the AOIs across a large environment introduces the following complexities: there is no fixed order for visiting the AOIs, and the UAV's objective is to identify as many EOIs as possible within the given time, which is insufficient to cover all AOIs.

To achieve this, \manoeuvername{} employs a hierarchical approach, dividing the task into three distinct modules: \textbf{S}\textit{election} (high-level planning), \textbf{Na}\textit{vigation} (mid-level planning), and \textbf{C}\textit{overage} (low-level planning).
The AOI \textit{Selection} module leverages the belief map to compute a high-level route between AOIs and to allocate the exploration time for each AOI. Based on the output from the \textit{Selection} module and other relevant environment information from the world model, the \textit{Navigation} module then performs the path planning to reach the selected AOI, and once there, the \textit{Coverage} module plans how to systematically search for EOIs in the area.

\subsubsection{AOI Selection}

\newif\iflongnames
\longnamestrue

\iflongnames
\newcommand{\AOI}{\mathit{AOI}}
\newcommand{\EOI}{\mathit{EOI}}

\newcommand{\allocatedTime}{\mathit{allocatedTime}}
\newcommand{\order}{\mathit{o}}
\newcommand{\leaveTime}{\mathit{leaveTime}}
\newcommand{\expectedEOIs}{\mathit{k}}

\else
\newcommand{\AOI}{\mathit{AOI}}
\newcommand{\EOI}{\mathit{EOI}}

\newcommand{\allocatedTime}{\mathbf{t}}
\newcommand{\order}{\mathbf{o}}
\newcommand{\leaveTime}{\mathbf{l}}
\newcommand{\expectedEOIs}{\mathbf{k}}
\fi

The AOI Selection module aims to determine the approximate optimal sequence of AOIs to explore and allocate  appropriate amounts of time for each. We model this  as a constraint optimization problem \cite{blackmore2011chance}, where the objective function maximizes the likelihood of detecting EOIs within the allocated time while minimizing travel time. This is done by considering the size of each AOI, the travel distances between them, the required exploration time and the probability of each AOI containing an EOI. 
We used the MiniZinc~\cite{minizinc} modeling language to model the problem, and the Chuffed~\cite{chuffed} solver to produce solutions.

\begin{table*}[t]
\centering
\begin{tabular}{ccc|cc|c|cccc}
\toprule[0.4mm]
\rowcolor{mygray} Planning & Perception & World & \multicolumn{2}{c|}{F1 Score ($\uparrow$)} & Success & \multicolumn{4}{c}{Search Time ($\downarrow$)} \\
\rowcolor{mygray} Component & Component & Model & Offline & Online & Rate ($\uparrow$) & 1st & 2nd & 3rd & 4th \\ \hline \hline
Fields2Cover~\cite{Mier_Fields2Cover_An_open-source_2023} & YOLO-World~\cite{cheng2024yolo} & \ding{55} & 04.40 & 11.40 & 29.58 & 78.30 & 137.57 & 179.79 & 256.37 \\
\manoeuvername{} & YOLO-World~\cite{cheng2024yolo} & \ding{55} & 06.17 & 08.24 & 31.67 & 74.02 & 93.64 & 96.93 & 120.09 \\  
Fields2Cover~\cite{Mier_Fields2Cover_An_open-source_2023} & \perceptionname{} & \ding{55} & 24.44 & 29.07 & 36.17 & 66.78 & 100.95 & 162.81 & 186.02\\  
Fields2Cover~\cite{Mier_Fields2Cover_An_open-source_2023} & \perceptionname{} & \ding{51} & 30.56 & 40.15 & 36.32 & 60.39 & 134.13 & 167.68 & 165.92 \\  
\manoeuvername{} & \perceptionname{} & \ding{55} & 41.27 & 50.29 & 40.27 & 51.72 & 80.73 & 133.57 & 153.79 \\  
\manoeuvername{} & \perceptionname{} & \ding{51} & \textbf{52.07} & \textbf{54.12} & \textbf{61.82} & 61.25 & 98.19 & 138.98 & 134.78\\  
\bottomrule[0.4mm]
\end{tabular}
\caption{The quantitative comparison between the proposed methods with baselines.}
\label{tab:quantitative}
\end{table*}

\begin{table*}[t]
\centering
\begin{tabular}{ccccc|cc|cc|cc}
\toprule[0.4mm]
\rowcolor{mygray} SoTA & Color/Type & 2D & 3D Projection & World & \multicolumn{2}{c|}{F1 Score ($\uparrow$)} & \multicolumn{2}{c|}{Precision ($\uparrow$)} & \multicolumn{2}{c}{Recall ($\uparrow$)} \\
\rowcolor{mygray} VFMs & Classifiers & tracking & on center of & Model & Offline & Online & Offline & Online & Offline & Online \\ \hline \hline
\ding{55} & Zeroshot & \ding{55} & 2D bbox & \ding{55} & 05.39 & 30.08 & 05.79 & 32.61 & 05.08 & 33.69\\ 
\ding{55} & Zeroshot & \ding{55} & 3D pointcloud & \ding{55} & 25.28 & 35.24 & 28.72 & 40.09 & 23.21 & 39.60\\ 
\ding{51} & Zeroshot & \ding{55} & 3D pointcloud & \ding{55} & 30.73 & 40.12 & 40.28 & 57.92 & 26.39 & 39.92\\ 
\ding{51} & Linear-probing & \ding{55} & 3D pointcloud & \ding{55} & 41.27 & 50.29 & 45.28 & 65.36 & 38.75 & 42.56\\ 
\ding{51} & Linear-probing & \ding{55} & 3D pointcloud & \ding{51} & 49.29 & 53.40 & 57.58 & 67.92 & 44.89 & 46.76\\
\ding{51} & Linear-probing & \ding{51} & 3D pointcloud & \ding{51} & \textbf{52.07} & \textbf{54.12} & \textbf{59.82} & \textbf{68.71} & \textbf{47.77} & \textbf{47.34} \\
\bottomrule[0.4mm]
\end{tabular}
\caption{The ablation study of \perceptionname{} component.}
\label{tab:perception_ablation}
\vspace{-2mm}
\end{table*}

\subsubsection{Navigation to AOI}
Once an AOI is selected, the Navigation module generates a plan for travelling to that area by constructing a visibility graph~\cite{oommen1987robot} using information regarding KOZs and voxel occupancy. Subsequently, it executes an A*~\cite{Hart1968} algorithm
to determine the optimal path, ensuring avoidance of both obstacles and KOZ.

\subsubsection{Area Coverage}
After reaching an AOI, the coverage module plans the low-level search for EOIs. 
It begins by converting the AOI into a grid, thus, representing the coverage task as the exploration of all accessible grid points. 
To achieve this we first create an open set of points to be visited. Then the module greedily finds the nearest non-visited point from the starting position, and uses the A* algorithm to navigate to that point while avoiding any obstacles or KOZ. Subsequently, the UAV navigates towards that point along the computed path, and removes visited points from the open set. Once the open set becomes empty, or all EOIs are found, the search concludes.
The belief map in the World Model 
will be updated based on the EOIs found in that AOI, and the updated belief map will be used by the \textit{Selection} module to select the next AOI.

\section{Experiments} 
\label{sec:eval}

\subsection{Implementation Details}
\label{sec:implementation_details}

We implemented our proposed system by integrating the GRiD, World Model, and SNaC components, and compared its performance against a framework built using state-of-the-art (SoTA) solutions for perception\footnote{ The name ``perception'' is used to represent the perception and grounding for simplicity.} and planning. This comparison highlights the contribution of our system to the specific problem. Additionally, we conducted several ablation studies for each component to assess the impact of each module and the specific features they contribute.

\noindent \textbf{Toolkit in \perceptionname{}.} For \perceptionname{}, we use SoTA \cite{liuGrounding2023} for grounding, linear probed CLIP~\cite{radfordLearning2021} for color/type classifiers, and OCSort~\cite{cao2023observation} as the 2D tracker to assign tracking IDs to targets. After generating the 2D bounding box for the detected target by the visual grounding model, we use EfficientSAM~\cite{xiong2023efficientsam} to obtain the pixel mask of the target. Using the depth sensor data, we compute the 3D coordinates of all pixels within the mask as a point cloud. The 3D location of the target is determined by averaging the points within this cloud, and then the 3D location is sent to the world model for further reasoning. In ablation studies, we follow HYDRA~\cite{keHYDRA2024} to use GLIP~\cite{li2022grounded} for grounding and XVLM~\cite{zeng2021multi} for zeroshot color/type classifiers, as the original VFMs. 

\noindent \textbf{Baseline.} The baseline system is composed of YOLO-World~\cite{cheng2024yolo} for the perception component and Fields2Cover~\cite{Mier_Fields2Cover_An_open-source_2023} for the planning component. These components were originally selected for the DARPA ANSR program, ensuring their relevance and suitability for our task. YOLO-World represents the state-of-the-art in vision-language models (VLMs), being known for its efficiency and high performance in 2D grounding tasks. Since YOLO-World does not provide estimated segmentation masks for EOIs to enhance their 3D localization, we compute their 3D coordinates by projecting the center of the 2D bounding box using the available depth data. On the planning side, Fields2Cover is a symbolic, model-based planner widely used for autonomous planning due to its robustness and efficiency in navigating complex environments.

\subsection{Metrics}

The primary criterion for evaluating the system's performance is the correct identification of EOIs (i.e., reported within 5 meters of the ground truth position). We define the success rate (SR) as $n/N$, where $N$ is the total number of EOIs and $n$ is the number of successfully detected EOIs. This differs slightly from previous work~\cite{duan2022survey} to account for multiple EOIs. The average SR across test scenarios is the main measure of overall performance.
This metric integrates the performance of both components: the planning component must generate paths that allow the camera to capture frames containing EOIs, while the perception and reasoning component must accurately identify EOIs within those frames.

To evaluate the perception, reasoning and grounding component independently, we use common metrics such as F1, Precision, and Recall. Prediction and ground truth targets are matched based on 3D coordinates. We compute these metrics for both \textbf{online} (frame-level) and \textbf{offline} (scenario-level) EOI reports. 
For an independent evaluation of the planning component, we use search time as a measure of navigation efficiency. Since this is only meaningful when approaches achieve the same SR,  we report search times for each target found (first, second, third, and fourth).

\begin{figure}[t]
    \centering
    \begin{subfigure}[t]{0.5\linewidth}
        \centering
        \includegraphics[width=0.85\linewidth]{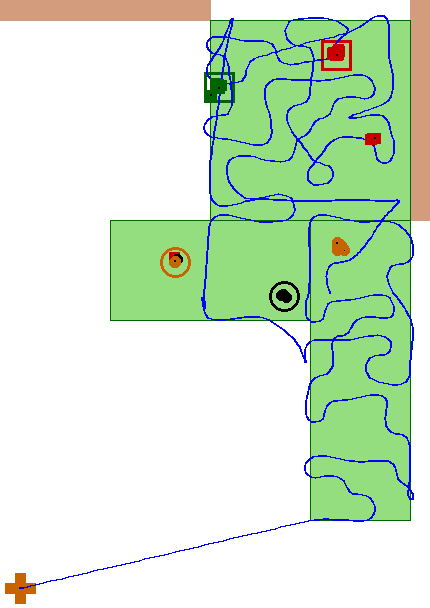}
        \caption{\methodname{}\\(\perceptionname{}, World Model, \manoeuvername{})}
    \end{subfigure}%
    ~ 
    \begin{subfigure}[t]{0.5\linewidth}
        \centering
        \includegraphics[width=0.85\textwidth]{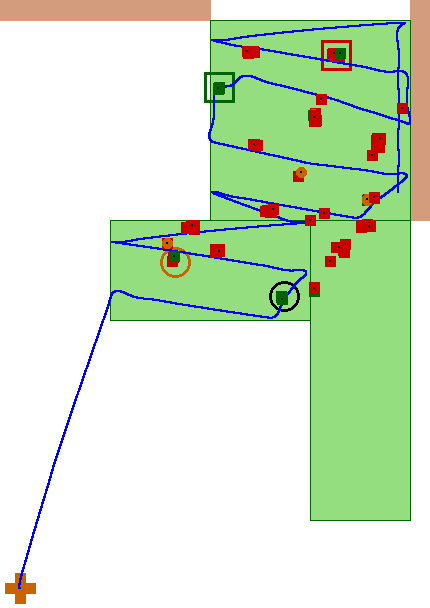}
        \caption{Baseline\\(YOLO-World, Field2Cover)}
    \end{subfigure}
    \caption{Comparison of (a) our proposed system and (b) the baseline method on the scenario depicted in Figure~\ref{fig:missionVisualisation}.
    Filled, colored shapes denote EOI reports, and blue curves represent the UAV's flight path.}
    \label{fig:qualitative}
    \vspace{-2mm}
\end{figure}

\subsection{Quantitative Comparison}

We compared the F1 score, SR (success rate), and search time across different configurations of planning and perception components, as shown in Table~\ref{tab:quantitative}.
When replacing Fields2Cover with \manoeuvername{}~(row 2), we observe a significant improvement in navigating efficiency, particularly when searching for multiple EOIs, and most notably when finding the 4th or final EOI. 
Similarly, comparing rows 1 and 3, where YOLO-World is replaced with \perceptionname{}, there is a dramatic improvement in EOI localization performance, as indicated by the F1 score. 
It is worth noting that noisy reports from YOLO-World lead to higher success rates~(around $30\%$) than would be expected, as only one report needs to be correct, and success rate does not adequately penalize incorrect reports.
The F1 Score metric gives a stronger indication of the actual performance of perception systems, and \perceptionname{} outperforms YOLO-World on this metric by around $20\%$.
Further, the combination of \perceptionname{} with \manoeuvername{} (row 5) leads to a substantial increase in mission F1 score, success rate, and search time.
The routes that \manoeuvername{} produces allow the \perceptionname{} and world model to see cars in the environment from more directions, allowing for higher confidence~(resp, lower confidence) to be built before making a report.
Finally, with the addition of the world model in rows 4 and 6, we see a further improvement, in particular in terms of the success rate and offline F1, demonstrating the effectiveness of our compositional neuro-symbolic approach.
Note that the slight increase in search time~(for 1st and 2nd targets) is due to the world model being more conservative, i.e., reporting only when confidence has reached a sufficient threshold.

\subsection{Qualitative Comparison}

To better understand the results of our experiments, we examined visualisations of the behaviour of the different configurations.
Figure~\ref{fig:qualitative} provides an example of flight paths and EOI reports from a) our proposed system \methodname{}, and b) the baseline system based on Field2Cover and YOLO-World.
From these figures, it can be seen that both approaches adequately cover the AOIs that contain EOIs.
The Fields2Cover approach employs a deliberate back and forth search strategy that systematically covers the AOIs.
However, YOLO-World produces many false positives~(filled red squares away from the targets), and also generated noisy output near the ground truth targets.
\methodname{}'s planning component \manoeuvername{} performs a more bespoke exploration that allows \perceptionname{} and the world model to see potential EOIs from more angles, thus making more confident  reports.
Overall, the qualitative visualisation shows the advantages of our integrated neuro-symbolic system in both navigation efficiency and target detection performance.

\subsection{Ablation Studies}

\noindent \textbf{\perceptionname{}.} We conducted extensive ablation studies to evaluate the impact of different modules in the \perceptionname{} component. Table~\ref{tab:perception_ablation} presents the results for various configurations, to evaluate the contributions of each module in \perceptionname{}. Online and offline perception metrics (F1, precision and recall) are used for comparison.
The first two rows highlight the impact of 3D projection methods, demonstrating that point cloud-based 3D projection significantly outperforms projecting the center point of 2D bounding boxes. The results from rows 2, 3, and 4 show the substantial positive contribution of SoTA VFMs and color/type classifiers. Finally, comparing rows 4, 5, and 6, we observe the effectiveness of integrating the world model and 2D tracking, both of which lead to notable performance improvements.

\begin{table}[h]
\centering
\scalebox{1}{\begin{tabular}{l|cc|c}
\toprule[0.4mm]
\rowcolor{mygray} World Model Component & \multicolumn{2}{c|}{F1 Score ($\uparrow$)} & Success \\
\rowcolor{mygray} Information Accumulation & Offline & Online & Rate ($\uparrow$) \\ \hline \hline
World reasoning only & 48.68 & 42.57 & 59.78 \\
+ Na\"ive Accumulation & 48.55 & 44.62 & 58.33 \\
+ Bayesian Filtering & \textbf{52.07} & \textbf{54.12} & \textbf{61.82} \\
\bottomrule[0.4mm]
\end{tabular}}
\caption{The ablation study of World Model.}
\label{tab:world_model_ablation}
\vspace{-2mm}
\end{table}

\noindent \textbf{World Model.} Ablation studies for the world model are presented in Table~\ref{tab:world_model_ablation}.
Starting with a basic version that only performs basic world reasoning, we see that the addition of information accumulation with na\"ive filtering~(using the average 3D position), only provides a small improvement for online F1 Score~($42.57\% \rightarrow 44.62\%$).
When Bayesian filtering is added we see a much larger improvement, in particular we see a $10\%$ increase in online F1 Score, demonstrating the importance of correctly handling uncertainty.

\begin{table}[h]
\centering
\scalebox{0.88}{\begin{tabular}{l|c|cccc}
\toprule[0.4mm]
\rowcolor{mygray} Planning & Success & \multicolumn{4}{c}{Search Time ($\downarrow$)} \\
\rowcolor{mygray} Component & Rate ($\uparrow$) & 1st & 2nd & 3rd & 4th \\ \hline \hline
Baseline & 20.83 & 96.29 & 104.64 & 70.14 & -  \\
+ AOI Selection & 43.75 & 71.17 & 133.20 & 156.69 & 210.41 \\
+ MiniZinc Optimization & 53.45 & 79.88 & 136.59 & 195.67 & 260.27  \\
+ Area coverage & \textbf{54.51} & 88.90 & 133.95 & 168.54 & 219.35 \\
\bottomrule[0.4mm]
\end{tabular}}
\caption{The ablation study of \manoeuvername{} component.}
\label{tab:maneuver_ablation}
\vspace{-2mm}
\end{table}

\noindent \textbf{\manoeuvername{}.} Table~\ref{tab:maneuver_ablation} presents the ablation study for the \manoeuvername{} component with ground truth perception which reports the targets in 25 meters. Starting with the baseline version, which employs Fields2Cover \cite{mier2023fields2cover} for area coverage and a randomly selected route for the AOIs, the introduction of AOI \textit{Selection} based on the belief map shows a significant improvement in the success rate, nearly doubling it (from 20.83\% to 43.75\%) while also reducing the search time. Incorporating MiniZinc optimization further enhances both the success rate (up to 53.45\%) and planning efficiency by providing a more optimized method for determining the AOI visitation order and exploration time allocation. Finally, the addition of the proposed area \textit{Coverage} module raises the success rate to 54.51\%, demonstrating its effectiveness in improving low-level search coverage throughout the environment.

\section{Conclusion}

This paper presented \methodname{}, a compositional neuro-symbolic system for autonomous UAVs in complex search missions. By integrating neuro-symbolic perception (\perceptionname{}), a probabilistic world model, and a hierarchical symbolic planning component (\manoeuvername{}), our approach enables efficient target detection, reasoning, and navigation. Extensive experiments demonstrate that \methodname{} significantly outperforms baselines in success rate, search efficiency, and localization performance.

\noindent \textbf{Broader Impact.} \methodname{} has potential for real-world applications such as search-and-rescue missions, improving UAVs' ability to locate targets in hazardous environments.

\noindent \textbf{Limitations.} Our system has been tested only in simulated environments and relies on accurate positional data, voxel grids, and point clouds. This can be addressed in future work.

\bibliography{main_arxiv}
\bibliographystyle{IEEEtranS}

\end{document}